\documentclass{igec2026}

\usepackage[utf8]{inputenc}
\usepackage{amsmath,amssymb}
\usepackage{graphicx}
\usepackage{url}
\usepackage[hidelinks]{hyperref}
\usepackage{algorithm}
\usepackage{algpseudocode}
\usepackage{bm}
\usepackage{multirow}
\usepackage{booktabs}
\usepackage{tabularx}
\usepackage{makecell}
\usepackage{pgfplots}
\pgfplotsset{compat=1.18}
\usepackage{colortbl}

\definecolor{futureColor}{HTML}{FF8C00}
\definecolor{pastColor}{HTML}{2E7BD4}
\definecolor{staticColor}{HTML}{696969}
\definecolor{sigGreen}{HTML}{C8E6C9}
\definecolor{sigRed}{HTML}{FFCDD2}

\papernumber{XXX}

\title{Short-Term Forecasting of Heterogeneous Wind Power Fleets: Global Temporal Fusion Transformer Leveraging Static Covariates}
\author{
  Viktor Walter$^{1}$, Andreas Wagner$^{1,2}$ \\
  \vspace{0.5em} 
  $^{1}$ Faculty of Management Science and Engineering, Karlsruhe University of Applied Sciences, Karlsruhe, Germany \\
  $^{2}$ Fraunhofer Institute for Industrial Mathematics, Kaiserslautern, Germany \\
  \vspace{0.5em}
  E-mail: viktor.walter@h-ka.de
}

\begin{document}

\maketitle

\begin{abstractsection}
With the increasing integration of renewable energy sources, accurate short-term wind power forecasting has become critical for grid stability and efficient electricity trading.
While deep learning, particularly the transformer architecture, has emerged as the state-of-the-art approach, current industry practices predominantly rely on specialized local models trained for individual assets.
This study investigates the efficacy of a global temporal fusion transformer (TFT) trained on a heterogeneous portfolio of 100 wind farms across Germany.
The primary objective is to determine whether a global model trained jointly on multiple wind farms can outperform local models trained exclusively on single sites in 48-hour-ahead forecasting.
Combining numerical weather prediction (NWP) data from the regional ICON-D2 and the global ECMWF HRES models, we compare the performance of a global TFT against distinct local models as well as a linear-regression baseline.
The global model leverages static context variables, such as farm age, altitude above sea level, turbine hub height, rotor diameter and power curve parameters to learn shared representations across diverse sites.
We evaluate on 100 synthetic wind farm profiles and, additionally, on a real operational portfolio of 11 German wind farms, across two seasonal test periods with intermediate retraining.
The empirical results demonstrate that the global TFT outperforms the local baselines for the majority of sites (87\% of the synthetic and 77\% of the real wind farms) while clearly surpassing the linear-regression baseline.
Statistical analysis using the Wilcoxon signed-rank test confirms the significance of these performance gains, which hold across both seasons and on the real portfolio.
Crucially, we show that this superiority is strictly conditional on the inclusion of static covariates.
A global model lacking asset-specific context fails to generalize effectively.
Furthermore, the interpretability of the TFT's variable selection network (VSN) yields interpretable insights into the learned feature selection.
The analysis identifies altitude above sea level as a relevant static discriminator on both synthetic and real data, while hub height emerges as a decisive feature on the real portfolio.
Furthermore, it reveals a dual modeling strategy for dynamic inputs: while the encoder chooses historical power as the most relevant observed feature, the decoder draws on a combination of upper-level wind speeds from both NWP models rather than a single atmospheric level or source, consistent with averaging out model-specific NWP errors.
These findings support a shift from single-site to scalable multi-site forecasting architectures, offering a pathway to reduce operational complexity while enhancing predictive accuracy in large-scale renewable energy portfolios.
\end{abstractsection}


\section{Introduction}
\label{sec:intro}

Renewable energy (RE) has established itself as a cornerstone of global energy infrastructure.
In 2024, RE accounted for 17\% of global electricity production, with wind energy contributing 25\% of this share \citep{statisticalreview2024}.
Wind power contributed 33\% to Germany's net electricity production \citep{fraunhoferise2025electricity}.
However, the inherent stochasticity of wind introduces significant fluctuations into the grid.
Consequently, market participants in short-term electricity trading, particularly on short-term markets such as day-ahead and intraday markets, rely heavily on accurate forecasts to mitigate volume risk.

Historically, forecasting methods have shifted from statistical approaches like ARIMA \citep{kavasseri2009day} to data-driven machine learning.
Hybrid neural networks have become the standard, effectively learning non-linear relationships between weather conditions and power generation \citep{guo2021short}.
Yet, current industry practice typically relies on site-specific models calibrated for individual wind farms.
These models often generalize poorly when data is scarce, a limitation stemming from the heterogeneous nature of the domain: wind farms differ significantly in both local meteorology and technical turbine parameters.
A global model can only generalize effectively if it learns to recognize these underlying distinctions.

A seminal architecture for such multi-horizon problems is the temporal fusion transformer (TFT) \citep{lim2021temporal}.
The TFT is uniquely advantageous as it incorporates static context features, allowing the network to differentiate between heterogeneous time series.
While TFTs represent the state of the art, existing literature on their application to wind power remains limited.
This study addresses open questions regarding how a global model must be configured to outperform specialized local models and clarifies which features are prioritized in this process.

The main contributions of this paper are summarized as follows:
\begin{itemize}
    \item We design a wind-speed--focused feature set that fuses two complementary NWP models---the high-resolution regional ICON-D2 by the German weather service (DWD) and the global European Centre for Medium-Range Weather Forecasts (ECMWF) highest resolution model (HRES)---across several atmospheric levels, supplying the model with complementary, height-resolved wind information.
    \item We show that the advantage of a global TFT over site-specific models is conditional on encoding physical static covariates: through a no-static ablation, a global model without asset-specific context fails to beat local baselines, whereas one with it does. We validate this finding on both synthetic profiles and a real operational portfolio of German wind farms, benchmarked against a linear-regression baseline.
    \item We leverage the explainable AI capabilities of the TFT's variable selection network (VSN) to elucidate the importance of various features that drive the heterogeneity between wind turbines, contrasting the feature attributions obtained on synthetic and real data.
\end{itemize}

The remainder of this paper is structured as follows:
The~\nameref{sec:related_work} section reviews relevant literature on wind power forecasting and TFT.
The~\nameref{sec:methodology} section details the feature selection and the model architecture.
The data, the hyperparameter optimization (HPO) and the evaluation is described in the \nameref{sec:experimental} section.
The~\nameref{sec:results} section presents and discusses the results of the experiments.
The last section concludes the paper and outlines directions for future work.
\section{Related Work}
\label{sec:related_work}

This section reviews existing literature pertinent to the present study.
Table~\ref{tab:literature_review} categorizes recent developments and positions the present study within this landscape.

Operational day-ahead forecasting fundamentally relies on NWP integration.
While approaches based solely on historical measurements \citep{kisvari2021wind, liu2024interpretable, mi2025physics} show promise for very short-term horizons (<6 hours), they are unsuitable for day-ahead scenarios where actual wind measurements are unavailable at the time of inference \citep{wang2017deep, tsai2023review}.
Consequently, recent advances predominantly focus on hybrid architectures fusing NWP data with deep learning \citep{xu2024multi, jachula2025wind}.

A critical distinction in current literature lies in the modeling scope.
Despite the shift toward transformers, the vast majority of research remains tethered to specialized, site-specific models \citep{verdone2025review}.
Only recently have global models emerged: \cite{cohen2025inverted} demonstrated that a global transformer trained on 74 sites could reduce errors by 33\% compared to local baselines, though this was limited to a 12-hour horizon and framed as post-calibration of NWP forecast errors on a power-capacity metric rather than direct power forecasting.

The TFT \citep{lim2021temporal} has become the architecture of choice for addressing heterogeneity in time series.
\cite{niu2024interpretable} and \cite{michalakopoulos2025hybrid} validated the TFT's ability to utilize static features to differentiate between sites.
However, a critical limitation in these studies is the scarcity of spatial diversity and the nature of the static context used.
While \cite{niu2024interpretable} incorporate static variables, their approach relies on categorical site identifiers (e.g., wind farm ID or season) rather than physical asset parameters.
Consequently, their model is conditioned to memorize specific site behaviors rather than learning transferable physical laws governed by turbine attributes.
Furthermore, their evaluation is restricted to only three wind farms.
Such a small sample size is likely insufficient for the model to learn a generalized mapping, as the variance across three samples prevents the disentanglement of site-specific noise from genuine physical characteristics.
Conversely, studies that achieve a higher site count, such as \cite{zhang2025fine} with 115 sites, typically ignore static covariates entirely (Table~\ref{tab:literature_review}).
Consequently, existing research either lacks the necessary scale to exploit static features effectively or neglects them altogether in favor of purely autoregressive or NWP-based patterns.

As summarized in Table~\ref{tab:literature_review}, a distinct research gap remains.
While components such as global training \citep{cohen2025inverted}, static feature integration \citep{wang2024very}, and day-ahead horizons \citep{cheng2024integrating} exist in isolation, no study has yet systematically combined a large-scale global TFT approach with NWP and static context for 48-hour power forecasting.
This work bridges this gap by experimentally comparing global versus local models on a dataset of 100 wind farms, ensuring sufficient heterogeneity to robustly quantify the impact of static features on forecast accuracy.

\begin{table}[h!]
    \caption{Literature Overview of Transformer-based Wind Power Forecasting}
    \label{tab:literature_review}
    \flushleft
    \begin{tabular}{lllllll}
        \toprule
        \textbf{Author} & \textbf{Model} & \textbf{NWP} & \textbf{Horizon} & \textbf{Scope} & \textbf{Sites} & \textbf{Static Cov.} \\
        \midrule
        \cite{mi2025physics} & TCN-TFT & - & 1h & Local & 1 & - \\
        \cite{liu2024interpretable} & TFT & - & 1h & Local & 1 & - \\
        \cite{hou2026short} & TFT & - & 1h & Local & 1 & - \\
        \cite{mo2024powerformer} & Transformer & - & 4 h & Local & 1 & - \\
        \cite{van2023temporal} & TFT & - & 24 h & Local & 1 & - \\
        \cite{michalakopoulos2025hybrid} & TFT & - & 24 h & Local & 2 & \checkmark \\
        \cite{wang2024very} & TFT & \checkmark & 4h & Local & 1 & \checkmark \\
        \cite{jachula2025wind} & TFT & \checkmark & 60 h & Local & 1 & - \\
        \cite{xu2024multi} & Transformer & \checkmark & 72 h & Local & 1 & - \\
        \midrule
        \cite{zhang2025fine} & TFT & \checkmark & 4 h & Global & 115 & - \\
        \cite{cohen2025inverted} & Transformer & \checkmark & 12 h & Global & 74 & - \\
        \cite{niu2024interpretable} & TFT & \checkmark & 24 h & Global & 3 & \checkmark \\
        \cite{cheng2024integrating} & Transformer & \checkmark & 48 h & Global & 7 & - \\
        \midrule
        \textbf{This Work} & TFT & \checkmark & 48 h & Global & 100 & \checkmark \\
    \bottomrule
    \end{tabular}
\end{table}

\section{Methodology}
\label{sec:methodology}

This section details the methodology, starting with the model architecture, followed by the feature selection.

\subsection{Temporal Fusion Transformer}
\label{subsec:tft}

The temporal fusion transformer (TFT) is an attention-based architecture for multi-horizon time series forecasting that captures multi-scale temporal dynamics while preserving interpretability through its variable selection network (VSN) \citep{lim2021temporal}.
The following description highlights the components most relevant to the wind energy application; we refer to the original publication for the complete framework.

\subsubsection{Input Processing and Feature Differentiation}
\label{subsubsec:tft_input}

A distinct advantage of the TFT is the separate handling of diverse input types relevant to wind power generation.
The model partitions inputs into static covariates, past-observed inputs, and known future inputs.
Static covariates encompass time-invariant attributes to learn from context.
These parameters condition the model to specific site characteristics.
Past-observed inputs utilize historical time series including active power generation.
Known future inputs integrate external information available at the time of prediction.
This specifically includes NWP data.
Known inputs span the entire sequence, covering both the observation window and the forecast horizon.

The experimental setup utilizes a 48-hour forecast horizon and a corresponding 48-hour lookback window, the horizon being limited to 48 hours by the NWP model.
Let $t$ denote the relative time index, with $t=0$ the moment of prediction.
The \textbf{past-observed inputs} $\mathbf{x}^{(obs)}$ are defined over the lookback window ($-47 \le t \le 0$), yielding $\lvert \mathbf{x}^{(obs)} \rvert = 48$, whereas the \textbf{known future inputs} $\mathbf{x}^{(known)}$ span both the lookback window and the forecast horizon ($-47 \le t \le 48$), yielding $\lvert \mathbf{x}^{(known)} \rvert = 96$.
Finally, \textbf{static covariates} $\mathbf{x}^{(static)}$ possess no temporal dimension and remain constant across all time steps.

\subsubsection{Variable Selection Network}
\label{subsubsec:tft_vsn_n_grn}

Wind power forecasting involves high-dimensional datasets due to characteristics of NWP models.
NWP models provide meteorological variables at multiple height levels and for many gridpoints close to the wind farm.
Turbine hub heights also vary significantly across different installation sites.
The TFT handles this complexity through its variable selection network.
The VSN assigns specific learnable weights to each input feature independently, which creates interpretability of the feature importances.
This weighting process occurs strictly before the data enters the main recurrent network components.
The mechanism effectively filters out irrelevant height levels or redundant sensor data, which can be highly site-specific.
This suppression drastically reduces the effective feature space.

Mathematically, this suppression represents a projection from a multi-feature space into a single latent representation via a convex combination.
Initially, the $m$ input features are transformed into distinct hidden state vectors of size $d_{model}$, creating an intermediate feature space in $\mathbb{R}^{m \times d_{model}}$.
The VSN aggregates these embedding vectors $\tilde{\xi}_t^{(j)}$, representing the $j$-th feature at time step $t$, using learnable selection weights $\nu_t^{(j)}$ to filter out irrelevant information through a weighted sum:
\begin{flalign}
& \tilde{\xi}_t = \sum_{j=1}^{m} \nu_t^{(j)} \tilde{\xi}_t^{(j)} &
\end{flalign}
Consequently, the effective feature dimension is drastically reduced, resulting in a single compact representation vector for the time step $t$ in $\mathbb{R}^{d_{model}}$.

\subsubsection{Static Covariate Encoders}
\label{subsubsec:tft_static}

Wind farms comprise turbines with varying physical properties.
The static covariate encoders mechanism integrates these time-invariant attributes into the forecasting logic.
Primary static inputs include hub height, altitude above sea level, rotor diameter, farm age and power curve parameters (cut-in, cut-out, rated wind speed), as these factors define the physical identity of a site.
The TFT processes these features using dedicated encoders to produce abstract context vectors in the hidden space.
These vectors condition the entire network by initializing the recurrent layers and influencing variable selection.
This architecture allows a single global model to distinguish between different turbine types and site characteristics.
It effectively adapts predictions to the specific physical constraints of each asset without retraining.

\subsubsection{Implementation and Loss Function}
\label{subsubsec:implementation}

The implementation adheres to the original TFT framework.
The model is trained with the quantile (pinball) loss evaluated at the $0.5$ quantile, which yields a single median point forecast; full probabilistic multi-quantile forecasting is left for future work.

The source code is publicly available\footnote{\url{https://github.com/vikwal/forecasting\_framework}}.

\subsection{Feature Selection}
\label{subsec:preprocessing}

This section details the feature set.
A central design choice is to restrict the inputs to wind speed---the physically dominant driver of power (Eq.~\ref{eq:power})---rather than the full meteorological state, and to draw it from two complementary NWP models at several atmospheric levels.
As outlined in the methodology, input variables are categorized into past-observed, known future, and static features.
Table \ref{tab:features} provides a comprehensive list of all utilized variables.

\begin{table}[h!]
    \caption{Features used for model training}
    \label{tab:features}
    \flushleft
    \begin{tabular}{lll}
        \toprule
        \textbf{Past Observed} & \textbf{Known Future (NWP)} & \textbf{Static} \\
        \midrule
        Power & ICON-D2 wind speed at 78m & Power curve (cut-in, cut-out, rated wind speed) \\
        & ICON-D2 wind speed at 127m & Hub height \\
        & ICON-D2 wind speed at 184m & Rotor diameter \\
        & HRES wind speed at 100m & Farm age at inference time \\
        & HRES wind speed at 200m & Surface altitude above sea level \\
    \bottomrule
    \end{tabular}
\end{table}

The dominant role of wind speed follows directly from the physical definition of wind power:

\begin{flalign}
\label{eq:power}
&\    P = \frac{1}{2} \cdot \rho \cdot A \cdot C_p \cdot v^3 &
\end{flalign}

where $\rho$ denotes the air density, $A$ the rotor swept area, $C_p$ the power coefficient, and $v$ the wind speed.
Because generation scales with the cube of wind speed, accurate wind-speed information at the rotor-relevant heights is by far the primary driver of forecast skill.

To this end, we fuse wind-speed forecasts from two complementary NWP models: the high-resolution, convection-permitting DWD's regional ICON-D2 model (2 km) and the global ECMWF HRES model (9 km).
Combining a regional and a global model exposes the network to complementary forecast-error structures, which the VSN can exploit when weighting the individual inputs.
Wind speed is provided at several atmospheric levels spanning the rotor-relevant height range (78--184~m for ICON-D2 and 100--200~m for HRES), since hub heights differ markedly across the heterogeneous portfolio.
Rather than aggregating these layers a priori, we expose the individual levels to the model and leave the selection of the relevant ones to the variable selection network (Section~\ref{subsec:feature_importance}), preserving the full vertical information of the wind profile.
\section{Experimental Setup}
\label{sec:experimental}

The first subsection introduces the utilized dataset.
Subsequently, the hyperparameter optimization (HPO) is detailed.
Finally, the performance metrics and statistical tests used to evaluate the results are defined.

\subsection{The Data}
\label{subsec:data}

This study utilizes a public dataset \citep{walter5642955synthetic} combining synthetic German wind farm profiles with real NWP data.
Wind-speed forecasts are drawn from DWD's ICON-D2 and ECMWF HRES.
The dataset comprises hourly power generation profiles for 160 distinct locations across Germany.
Each site includes time series for six different turbine types and meteorological features consisting of both station measurements and NWP data.
The forecast data incorporates the 06:00, 09:00, 12:00, and 15:00 UTC ICON-D2 runs, each providing a 48-hour prediction horizon.
The data spans the period from July 2023 to Jan 2026 and is publicly available on Mendeley Data \citep{Walter2025Mendeley}.

\begin{figure}[H]
\flushleft
\includegraphics[width=0.8\linewidth]{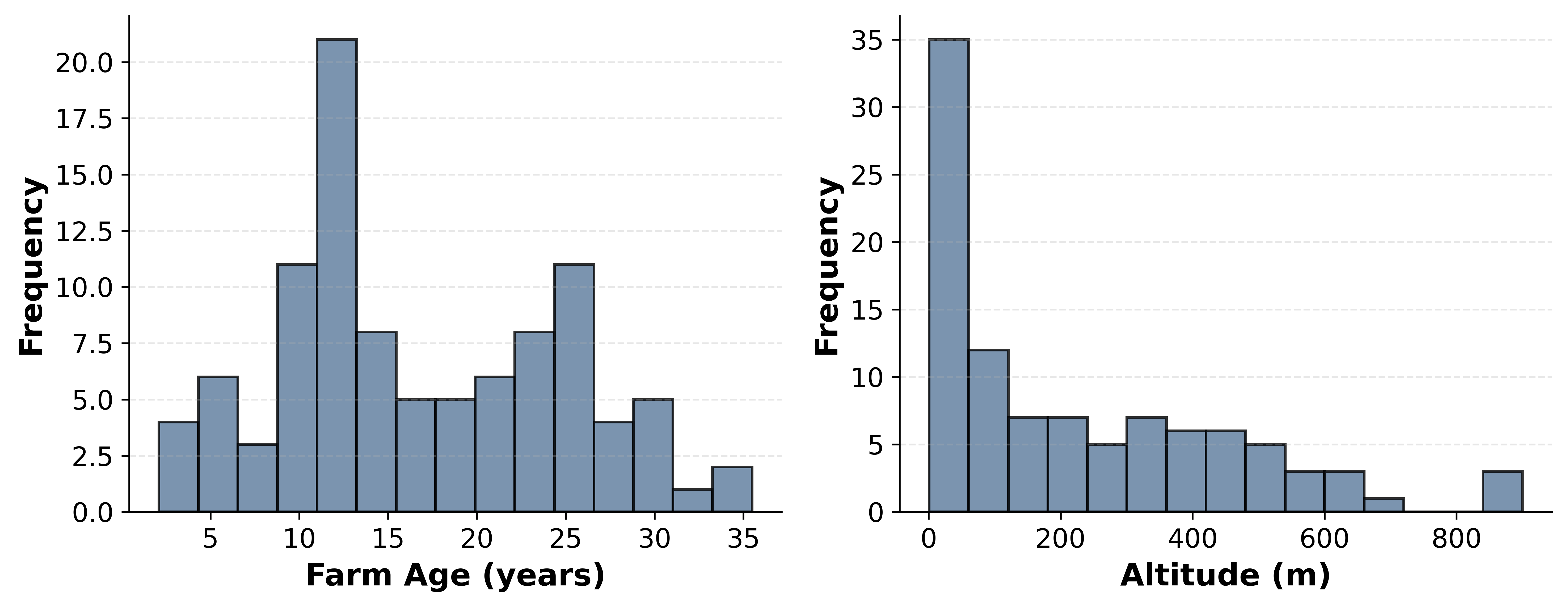}
\caption{Distribution of farm age and surface altitude above sea level}
\label{fig:fig1}
\end{figure}

For this study, 100 randomly distributed sites comprising wind turbines across Germany were selected.
To quantify the effect of portfolio size, we additionally define a fixed 50-farm subset (a random half of the 100 sites): a global model trained only on these 50 farms (\emph{Global-50}) is compared against the 100-farm model evaluated on the same subset, with both optimizing their hyperparameters.
Six synthetic turbine models recur throughout the sample.
Table~\ref{tab:turbines} lists their technical parameters and the share of wind farms featuring these turbines.
Figure~\ref{fig:fig1} illustrates the distribution of farm age and altitude above sea level.
The portfolio comprises both young and mature turbines.
Most wind farms are situated at low altitudes, though the range extends to 900 meters.
Here, altitude serves mainly as a proxy for more complex, elevated terrain, which affects forecasting difficulty.

\begin{table}[h!]
    \caption{Synthetic wind power turbines occurring in this study}
    \label{tab:turbines}
    \flushleft
    \begin{tabular}{llllllll}
        \toprule
        \textbf{Turbine} & \textbf{Rated Power} & \textbf{Hub} & \textbf{Diameter} & \textbf{Cut-in} & \textbf{Cut-out} & \textbf{Rated Speed} & \textbf{Frequency} \\
        \midrule
        Vestas V112     & 3.45 MW  & 119 m  &  112 m  & 3 m/s  & 25 m/s & 13 m/s  & 21 \\
        Enercon E-82 E2 & 2.0 MW   & 138 m  &  82 m   & 2 m/s  & 34 m/s & 13 m/s  & 17 \\
        Enercon E-115   & 2.5 MW   & 149 m  &  115 m  & 3 m/s  & 34 m/s & 12 m/s  & 16 \\
        Vestas V80      & 1.8  MW  & 78 m   &  80 m   & 4 m/s  & 25 m/s & 15 m/s  & 16 \\
        Enercon E-70 E4 & 2.3 MW   & 57 m   &  71 m   & 3 m/s  & 34 m/s & 15 m/s  & 15 \\
        Vestas V90      & 2.0 MW   & 95 m   &  90 m   & 4 m/s  & 25 m/s & 13 m/s  & 15 \\
    \bottomrule
    \end{tabular}
\end{table}

To validate the findings beyond synthetic profiles, we additionally evaluate the models on a real operational portfolio of 11 German wind farms distributed across the country.
The portfolio comprises 53 turbines with a total installed capacity of 120.45~MW and covers approximately one year of measured power generation.
In contrast to the synthetic profiles, this data is used without any prior cleaning and therefore reflects real operating conditions, including curtailment, maintenance, outages, and grid constraints.
This deliberately conservative setting allows us to assess whether the conclusions drawn from the synthetic experiments transfer to real-world SCADA behavior.
For the real portfolio, we reuse the hyperparameters obtained from the synthetic HPO without any re-tuning. This is a deliberate choice: it avoids information leakage from the real test periods (which precede the synthetic HPO validation window) and additionally tests the transferability of the configuration.

\subsection{Hyperparameter Optimization}
\label{subsec:hpo}

Hyperparameter optimization (HPO) utilizes a 12-fold rolling time series cross-validation.
The minimum training period (July 2023--July 2024) is followed by a validation period from August 2024 to July 2025.

In each iteration $k$, the model trains on all historical data up to the $k$-th validation month, progressively expanding the training set.
Optuna's tree-structured parzen estimator (TPE) is employed to minimize the average $\mathrm{RMSE}$ across all folds.

To accelerate the search, median pruning terminates unpromising trials early if intermediate performance (checked after the second fold) exceeds the median of prior trials.

\subsection{Evaluation}
\label{subsec:eval}

Following HPO, the models are retrained using the best-found hyperparameter set and evaluated on consecutive held-out test windows.
To probe seasonal generalization, the synthetic portfolio is assessed on two test periods: a first window covering August to October 2025 (late summer and early autumn) and a second window covering November 2025 to January 2026 (winter), with the models retrained on all data preceding each window.
The real-world portfolio is evaluated analogously on two consecutive three-month windows, the first under winter and the second under spring conditions, again with intermediate retraining.
This rolling protocol mirrors operational practice and tests whether the performance gains persist across seasons rather than being specific to a single period.

In addition to the local TFT baselines, we benchmark all configurations against a simple reference model: a linear regression on the NWP wind-speed inputs (LinReg).
This contextualizes the marginal value of the TFT architecture itself relative to a classical statistical predictor.

The predictive performance is assessed using the coefficient of determination ($\mathrm{R^2}$), root mean squared error ($\mathrm{RMSE}$), and mean absolute error ($\mathrm{MAE}$).
To ensure comparability across sites with varying capacities, the error metrics are normalized by the rated power and reported as percentages.
Given the ubiquity of these metrics in regression analysis, their formal mathematical definitions are omitted for the sake of brevity.

To statistically evaluate the performance differences between models, we employ the two-sided Wilcoxon signed-rank test on the paired per-site errors.
As a non-parametric paired test, it makes no assumption of normality for the differences, which makes it robust for the limited and potentially skewed sample sizes encountered here.
We treat a difference as significant at the 5\% level ($p<0.05$); to guard against false positives from multiple comparisons, all reported significant results additionally survive Holm--Bonferroni correction across the five tests.

\section{Results}
\label{sec:results}

This section presents the empirical results of the study.
First, Figure~\ref{fig:fig2} compares the global and the locally trained models on both the synthetic and the real portfolio, with and without static covariates.
The corresponding Wilcoxon signed-rank tests are summarized in Table~\ref{tab:significance}.
Subsequently, the feature importance for future, past, and static covariates is depicted in Figure~\ref{fig:fig3}.
Tables~\ref{tab:test_performance} and~\ref{tab:test_performance_real} summarize the performance metrics for the synthetic and the real portfolio across all model configurations and test periods.
Finally, Table~\ref{tab:hyperparameters} lists the optimal hyperparameters derived from the validation data.

\subsection{Global vs. Local Models}
\label{subsec:global_vs_local}

\begin{figure}[H]
\flushleft
\includegraphics[width=1\linewidth]{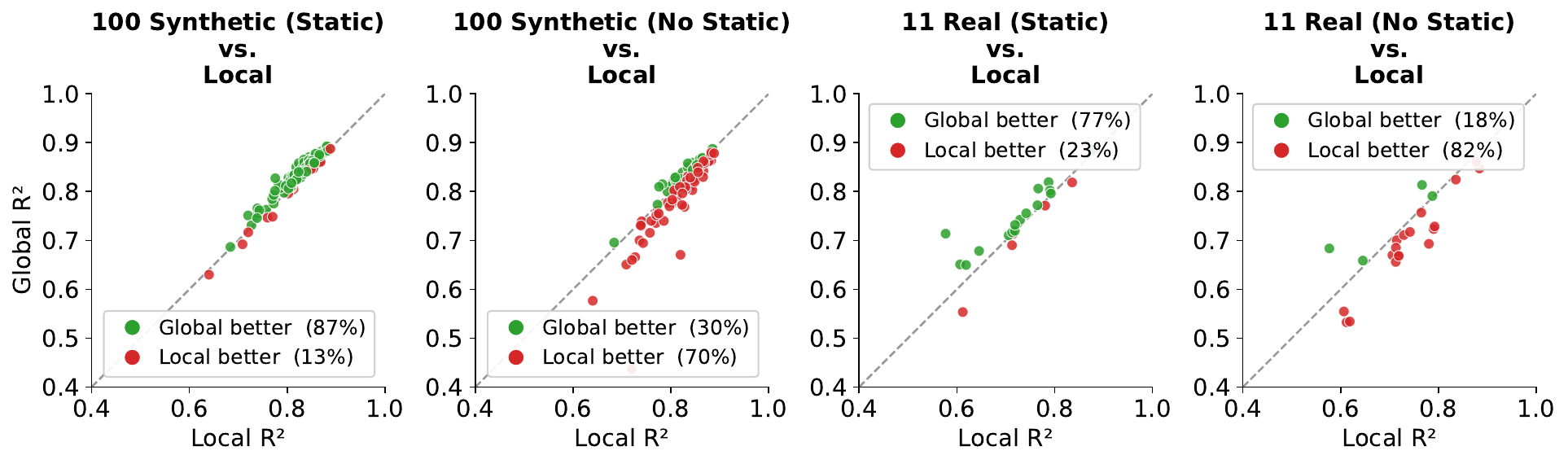}
\caption{Per-farm $\mathrm{R^2}$ of the global model against the local baseline, for the synthetic (100 farms) and the real (11 farms) portfolios, each with and without static covariates. Points above the diagonal (green) favor the global model.}
\label{fig:fig2}
\end{figure}

Figure~\ref{fig:fig2} shows four scatterplots of the per-farm test-set $\mathrm{R^2}$ of the global model (y-axis) against the local models (x-axis); points above the diagonal (green) favor the global model, those below (red) the local model.

When static covariates are included, the global model outperforms the local baseline for the clear majority of sites: 87\% of the synthetic farms and 77\% of the real farms lie above the diagonal.
Removing the static context reverses this picture entirely.
Without it, only 30\% of the synthetic and 18\% of the real farms still benefit from global training, i.e.\ the local models prevail for the majority of sites.
This pattern is remarkably consistent between the synthetic and the real portfolio.

Table~\ref{tab:significance} confirms these observations statistically.
Using a two-sided Wilcoxon signed-rank test, the static global model significantly improves on the local baseline on the synthetic data (both the 100- and the 50-farm configuration) and on the real portfolio.
Conversely, both no-static configurations are significantly \emph{worse} than the local baseline, underscoring that without static covariates global training actively degrades accuracy.
All five comparisons remain significant after Holm--Bonferroni correction.
This corroborates the key finding: static covariates are a prerequisite for a global model to reliably outperform site-specific training, and this requirement holds on real-world data just as on the synthetic profiles.

\begin{table}[h!]
    \caption{Two-sided Wilcoxon signed-rank test on $\mathrm{RMSE}$ against the local baseline. Green/red columns denote a statistically significant improvement/degradation of the global model relative to the local baseline ($p<0.05$); all five comparisons remain significant after Holm--Bonferroni correction. ($^{*}p<0.05$, $^{**}p<0.01$, $^{***}p<0.001$.)}
    \label{tab:significance}
    \flushleft
    \begin{tabular}{l >{\columncolor{sigGreen}}c >{\columncolor{sigRed}}c >{\columncolor{sigGreen}}c >{\columncolor{sigGreen}}c >{\columncolor{sigRed}}c}
        \toprule
         & \multicolumn{3}{c}{\textbf{Synthetic}} & \multicolumn{2}{c}{\textbf{Real}} \\
        \cmidrule(lr){2-4} \cmidrule(lr){5-6}
         & 100 (static) & 100 (no static) & 50 (static) & Real (static) & Real (no static) \\
        \midrule
        Median rel. $\Delta\mathrm{RMSE}$ [\%] & $+2.42$ & $-3.04$ & $+2.01$ & $+1.64$ & $-6.19$ \\
        $p$-value & $2.6\!\times\!10^{-10}$ & $1.8\!\times\!10^{-7}$ & $8.9\!\times\!10^{-6}$ & $0.039$ & $0.0074$ \\
        Significance & $^{***}$ & $^{***}$ & $^{***}$ & $^{*}$ & $^{**}$ \\
        \bottomrule
    \end{tabular}
\end{table}

The 50-farm configuration additionally addresses the question of scale.
Training the same 50 farms as part of the larger 100-farm portfolio yields comparable, but marginally higher accuracy (Table \ref{tab:test_performance}), showing that scaling the portfolio does not dilute per-site performance.

\subsection{Feature Importance}
\label{subsec:feature_importance}

\begin{figure}[h!]
\flushleft
\includegraphics[width=1\linewidth]{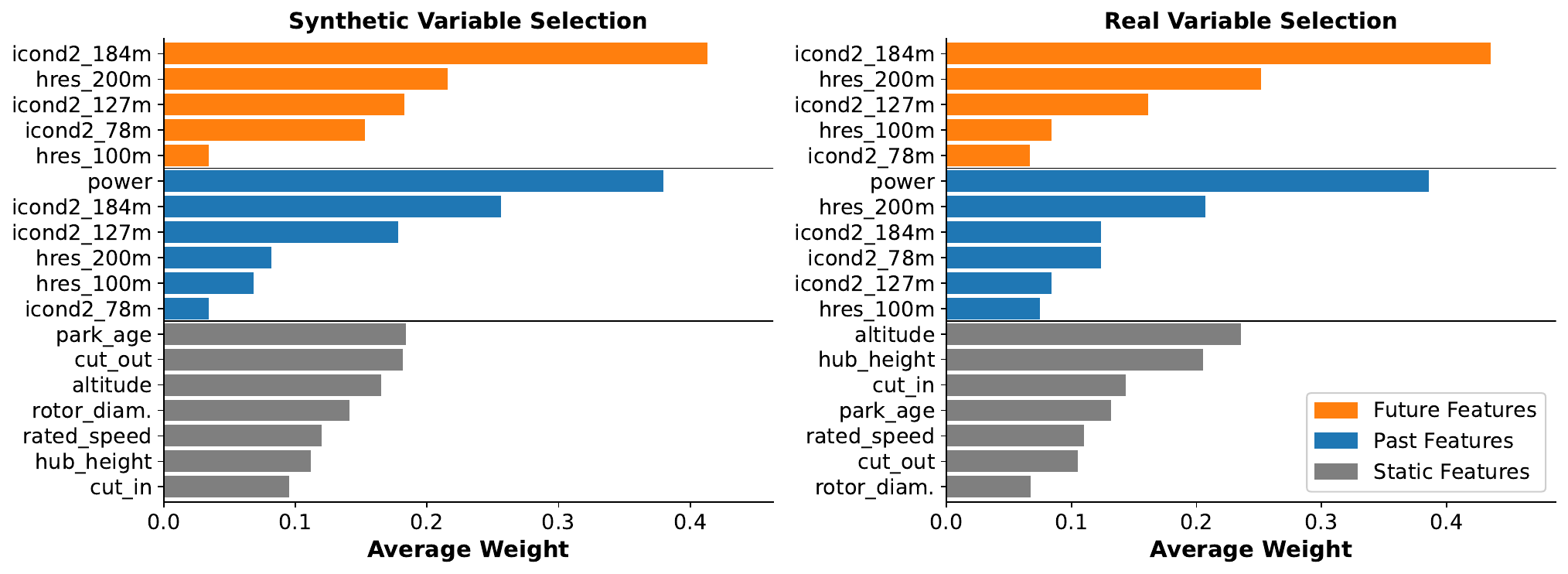}
\caption{Feature importance weights of the VSN for the model trained on synthetic (left) and real (right) data.}
\label{fig:fig3}
\end{figure}

The VSN inherent to the TFT architecture enables the interpretability of feature importance.
Figure~\ref{fig:fig3} displays the average importance weights for future, past, and static covariates, separately for the model trained on synthetic (left) and on real (right) data.

Remarkably, the three most important future features are identical on both datasets: the ICON-D2 wind speed at 184~m, the HRES wind speed at 200~m, and the ICON-D2 wind speed at 127~m.
The decoder thus distributes its weight across several upper-level wind speeds drawn from both NWP models rather than relying on a single layer or a single source, which is consistent with a strategy of averaging out the systematic forecast errors of the individual models.
For the past (observed) inputs, historical power output is the most important predictor on both datasets, complemented by the same upper-level wind speeds.

The static features reveal the most instructive differences between the synthetic and the real data.
Altitude above sea level is assigned a consistently high weight of roughly 20\% in both cases, making it the most robust static discriminator between sites.
This is physically plausible, as altitude acts as a proxy for complex or elevated terrain, which complicates the interdependencies between the NWP layers.
Farm age, by contrast, receives a high weight on the synthetic data but a markedly lower one on the real portfolio.
This is expected: in the synthetic profiles, turbine ageing is modeled explicitly, which artificially inflates the relevance of this feature, whereas its effect is weaker and noisier in reality.
Research nonetheless indicates that real turbines aged 20 years or older may experience up to a 35\% decline in capacity factor \citep{staffellHowDoesWind2014}.
The comparatively high weight of the cut-out speed on the synthetic data is likewise unexpected, since cut-out events (25--34~m/s) are physically rare; we therefore attribute it to an artifact of the synthetic generation process rather than to a genuine physical driver.

Most notably, hub height emerges as a decisive static feature on the real data, ranking second, while it plays only a minor role on the synthetic data.
This is plausible: the hub height determines which atmospheric levels are physically relevant for a given turbine and therefore directly conditions the selection of the NWP wind-speed inputs.
The comparatively weak role of hub height on the synthetic data likely stems from the wind-speed extrapolation method used to generate the synthetic profiles, which does not reproduce this height dependence one-to-one.
This contrast offers valuable context to the prior work of \cite{michalakopoulos2025hybrid}: on real measurements, hub height is in fact among the most informative static covariates.

\subsection{Performance on Test Data}
\label{subsec:performance}

Tables~\ref{tab:test_performance} and~\ref{tab:test_performance_real} report the normalized performance metrics for all configurations across both test periods, separately for the synthetic (100- and 50-farm sets) and the real portfolio.
The static global model achieves the best accuracy throughout, reaching $\mathrm{R^2} = 86.2\%$, $\mathrm{RMSE} = 8.0\%$ and $\mathrm{MAE} = 4.7\%$ on the synthetic late-summer and early-autumn window, and remaining the strongest model in the more challenging winter period.
Performance differences between the 100-farm and the 50-farm set are small, but the 100-farm model is marginally better, indicating that it gains per-site accuracy due to the increased training sample size.

Two observations stand out.
First, the linear regression on NWP inputs reaches $\mathrm{R^2} \approx 61$--$71\%$, well below the TFT models, confirming that the task is non-trivial and that the marginal value of the TFT architecture itself is substantial relative to a classical statistical predictor.
Second, and most importantly, the results on the real portfolio confirm the synthetic findings: the static global model again outperforms both the local baseline and the no-static variant in both seasons, despite the raw, uncleaned nature of the real SCADA data, although in the spring window the margin over the local model is small ($72.7$ vs.\ $72.1$ in $\mathrm{R^2}$).

It should be emphasized that the real-world data was used without any prior cleaning.
The signal therefore contains curtailment, maintenance shutdowns, outages, and grid-induced deviations that are physically unpredictable from NWP inputs alone and impose a hard ceiling on the attainable accuracy.
Filtering such artifacts, for instance via density-based clustering such as DBSCAN, together with a longer training history and accurate turbine master data, can be expected to improve the achievable performance further.
That the global model already prevails under these conservative conditions underlines the robustness of the approach.

\begin{table}[h!]
    \caption{Normalized performance on the synthetic portfolio across both test periods, for the 100- and 50-farm evaluation sets. Best $\mathrm{R^2}$ per column in bold; ``--'' marks configurations not evaluated on the respective set.}
    \label{tab:test_performance}
    \flushleft
    \footnotesize
    \setlength{\tabcolsep}{3pt}
    \begin{tabular}{l*{12}{c}}
        \toprule
         & \multicolumn{3}{c}{\textbf{Test Period 1 (100 farms)}} & \multicolumn{3}{c}{\textbf{Test Period 2 (100 farms)}} & \multicolumn{3}{c}{\textbf{Test Period 1 (50 farms)}} & \multicolumn{3}{c}{\textbf{Test Period 2 (50 farms)}} \\
         & \multicolumn{3}{c}{Aug--Oct 2025} & \multicolumn{3}{c}{Nov 2025--Jan 2026} & \multicolumn{3}{c}{Aug--Oct 2025} & \multicolumn{3}{c}{Nov 2025--Jan 2026} \\
        \cmidrule(lr){2-4} \cmidrule(lr){5-7} \cmidrule(lr){8-10} \cmidrule(lr){11-13}
        \textbf{Model} & $\mathrm{R^2}$ & $\mathrm{RMSE}$ & $\mathrm{MAE}$ & $\mathrm{R^2}$ & $\mathrm{RMSE}$ & $\mathrm{MAE}$ & $\mathrm{R^2}$ & $\mathrm{RMSE}$ & $\mathrm{MAE}$ & $\mathrm{R^2}$ & $\mathrm{RMSE}$ & $\mathrm{MAE}$ \\
        \midrule
        Global-100 (static)    & \textbf{86.2} & 8.0 & 4.7 & \textbf{79.6} & 17.7 & 10.7 & \textbf{85.8} & 8.3 & 4.9 & \textbf{79.9} & 18.0 & 11.1 \\
        Global-100 (no static) & 84.2 & 8.4 & 5.0 & 76.6 & 18.6 & 11.3 & 84.1 & 8.7 & 5.2 & 77.2 & 19.0 & 11.7 \\
        Global-50 (static)     & -- & -- & -- & -- & -- & -- & 85.5 & 8.3 & 5.1 & 79.8 & 18.1 & 11.2 \\
        Local                  & 85.2 & 8.3 & 5.1 & 78.7 & 18.0 & 11.3 & 84.7 & 8.6 & 5.3 & 79.1 & 18.3 & 11.5 \\
        LinReg                 & 71.2 & 11.4 & 8.5 & 61.4 & 23.7 & 18.1 & 71.2 & 11.7 & 8.8 & 62.5 & 24.3 & 18.7 \\
        \bottomrule
    \end{tabular}
\end{table}

\begin{table}[h!]
    \caption{Normalized performance on the real portfolio (11 farms) across both test periods. Best $\mathrm{R^2}$ per period in bold.}
    \label{tab:test_performance_real}
    \flushleft
    \begin{tabular}{lcccccc}
        \toprule
         & \multicolumn{3}{c}{\textbf{Test Period 1}} & \multicolumn{3}{c}{\textbf{Test Period 2}} \\
         & \multicolumn{3}{c}{Dec 2023--Feb 2024} & \multicolumn{3}{c}{Mar--May 2024} \\
        \cmidrule(lr){2-4} \cmidrule(lr){5-7}
        \textbf{Model} & $\mathrm{R^2}$ & $\mathrm{RMSE}$ & $\mathrm{MAE}$ & $\mathrm{R^2}$ & $\mathrm{RMSE}$ & $\mathrm{MAE}$ \\
        \midrule
        Global (static)     & \textbf{76.2} & 15.3 & 10.7 & \textbf{72.7} & 11.6 & 7.7 \\
        Global (no static)  & 72.6 & 16.6 & 12.2 & 68.1 & 12.5 & 8.3 \\
        Local               & 74.2 & 16.2 & 11.5 & 72.1 & 11.8 & 8.0 \\
        LinReg              & 63.5 & 19.2 & 14.3 & 62.2 & 13.6 & 10.3 \\
        \bottomrule
    \end{tabular}
\end{table}

Table~\ref{tab:hyperparameters} reports the optimal hyperparameters of the best cross-validated model, while Table~\ref{tab:trainingcost} lists the training cost of each configuration.
Although the local models can be trained quickly in parallel, the global approach consolidates the entire fleet into a single model trained in a few GPU-hours, drastically reducing the number of models to maintain and deploy.

\begin{table}[h!]
    \caption{Optimal hyperparameters of the 100-farm model.}
    \label{tab:hyperparameters}
    \flushleft
    \begin{tabular}{ll}
        \toprule
        \textbf{Hyperparameter} & \textbf{Value} \\
        \midrule
        batch\_size & 46 \\
        learning\_rate & $1.95 \times 10^{-4}$ \\
        n\_heads & 8 \\
        hidden\_dim & 40 \\
        dropout & 0.0992 \\
        num\_lstm\_layers & 1 \\
        static\_embed\_dim & 128 \\
        clipnorm & 6.9179 \\
        \bottomrule
    \end{tabular}
\end{table}

\begin{table}[h!]
    \caption{Training cost per configuration on a single RTX A6000, split by test period ($\mathrm{TP}_1$, $\mathrm{TP}_2$). Each sample comprises a known-input window of $96\times5$, an observed window of $48\times1$, 7 static features, and a 48-step target. For the local models, $N$ is the per-site training-set size (the data is partitioned across the sites), whereas the listed duration is the total wall-clock for all sites trained concurrently on one GPU (4 parallel runs at 0.25~GPU each for the synthetic and 10 at 0.1~GPU each for the real data); all other configurations use a full GPU.}
    \label{tab:trainingcost}
    \flushleft
    \setlength{\tabcolsep}{6pt}
    \begin{tabular}{llcccc}
        \toprule
        \textbf{Data} & \textbf{Model} & \textbf{Duration}$_{\mathrm{TP}_1}$ & \textbf{Duration}$_{\mathrm{TP}_2}$ & $N_{\mathrm{TP}_1}$ & $N_{\mathrm{TP}_2}$ \\
        \midrule
        \multirow{4}{*}{Synthetic}
          & 100 static          & 2\,h\,12\,m & 1\,h\,46\,m & 292k & 329k \\
          & 100 no static       & 2\,h\,12\,m & 1\,h\,46\,m & 292k & 329k \\
          & 50 static           & 55\,m       & 1\,h\,13\,m & 146k & 164k \\
          & Local ($\times$100) & 42\,m       & 42\,m       & $\approx$2.9k & $\approx$3.3k \\
        \midrule
        \multirow{3}{*}{Real}
          & Global (static)     & 1\,m\,23\,s & 4\,m\,13\,s & 5.3k & 9.3k \\
          & Global (no static)  & 1\,m\,16\,s & 3\,m\,47\,s & 5.3k & 9.3k \\
          & Local ($\times$11)  & 1\,m\,24\,s & 1\,m\,29\,s & $\approx$0.5k & $\approx$0.8k \\
        \bottomrule
    \end{tabular}
\end{table}

\section{Conclusion}
\label{sec:conclusion}

This study evaluated the efficacy of a global TFT for 48-hour-ahead wind power forecasting across a heterogeneous portfolio.
The findings demonstrate that a global model can significantly outperform specialized local models, supporting a shift from single-asset to scalable multi-asset architectures.

The primary empirical result indicates that the global model outperforms the local baselines for the large majority of wind farms (87\% on the synthetic and 77\% on the real portfolio), and clearly surpasses the linear-regression baseline.
These gains are statistically significant and, crucially, persist across two seasonal test periods and on a real operational portfolio of 11 German wind farms.
However, this performance gain is critically dependent on the integration of static context variables.
It was demonstrated that a global model lacking asset-specific metadata fails to generalize significantly better than local benchmarks.
This confirms that explicitly modeling heterogeneity via parameters such as altitude and hub height is a prerequisite for successful cross-learning between sites.

Furthermore, the interpretability provided by the TFT's VSN offers valuable physical insights.
The feature importance analysis suggests the model employs a dual strategy: while the encoder utilizes historical power to physically reconstruct turbine behavior, the decoder combines upper-level wind speeds from both NWP models rather than relying on a single layer or source, consistent with averaging out model-specific NWP errors.
Notably, the three most influential future features are identical on the synthetic and the real data, whereas the static attributions differ in an interpretable way, with hub height emerging as a decisive discriminator on real measurements.

In summary, this work illustrates that global transformer models not only reduce operational complexity by consolidating model architectures but also measurably enhance forecast accuracy for the renewable energy sector through the utilization of fleet-wide data.
Future research should examine the transferability of these findings to probabilistic forecasting, assess the impact of systematic SCADA data cleaning, and explore the potential of integrating further NWP models and grid points to enhance forecast accuracy.

\section*{ACKNOWLEDGEMENT}
This work was supported by the German Federal Ministry of Research, Technology and Space [grant 13FH587KX1] (FederatedForecasts).

\vspace{12pt}
\noindent
\textbf{NOMENCLATURE}

\vspace{4pt}

{\setlength{\tabcolsep}{4pt}
\noindent
\begin{tabular}{@{}ll@{}}
    $A$ & Rotor swept area, $\mathrm{m^2}$ \\
    $C_p$ & Wind power coefficient \\
    $d_{model}$ & Latent (hidden) space dimension \\
    $j$ & Feature index \\
    $k$ & Iteration in cross validation \\
    $\mathrm{MAE}$ & Mean absolute error normalized by installed capacity \\
    $m$ & Number of features \\
    $P$ & Power, $\mathrm{W}$ \\
    $\mathrm{RMSE}$ & Root mean squared error normalized by installed capacity \\
    $\mathrm{R^2}$ & Coefficient of determination ($R^2$) \\
    $t$ & Time index \\
    $v$ & Wind speed, $\mathrm{m\,s^{-1}}$ \\
    $\lvert \mathbf{x}^{(known)} \rvert$ & Length of known input vector \\
    $\lvert \mathbf{x}^{(obs)} \rvert$ & Length of observed input vector \\
    $\nu_t^{(j)}$ & Selection weight of feature $j$ at $t$ \\
    $\xi_t^{(j)}$ & Processed embedding vector of feature $j$ at $t$ \\
    $\rho$ & Air density, $\mathrm{kg\,m^{-3}}$ \\
\end{tabular}
}

\bibliography{bibliography}

\end{document}